\documentclass[11pt,a4paper]{article}
\usepackage{authblk}
\usepackage[hyperref]{emnlp2020}
\usepackage{times}
\usepackage{latexsym}
\usepackage{subfigure}

\usepackage{color}
\usepackage{times}
\usepackage{latexsym}
\usepackage{graphicx}
\usepackage{float}
\usepackage{url}
\usepackage{booktabs}
\usepackage{dblfloatfix}
\usepackage{amsmath}
\usepackage{bbm}
\usepackage{siunitx}
\usepackage{mathptmx}
\usepackage{hyperref}

\aclfinalcopy 

\title{Contextualized moral inference}

\author[ ]{Jing Yi Xie}
\author[ ]{Graeme Hirst}
\author[ ]{Yang Xu}
\affil[ ]{Department of Computer Science, University of Toronto, Toronto, Canada}
\affil[ ]{\texttt {jingyi.xie@mail.utoronto.ca, \{gh,yangxu\}@cs.toronto.edu}}

\date{}

\begin{document}
\maketitle
\begin{abstract}
Developing moral awareness in intelligent systems has shifted from a topic of philosophical inquiry to a critical and practical issue in artificial intelligence over the past decades. However, automated inference of everyday moral situations remains an under-explored problem. We present a text-based approach that predicts people’s intuitive judgment of moral vignettes. Our methodology builds on recent work in contextualized language models and textual inference of moral sentiment. We show that a contextualized representation  offers a substantial advantage over alternative representations based on word embeddings and emotion sentiment in inferring human moral judgment, evaluated and reflected in three independent datasets from moral psychology. We discuss the promise and limitations of our approach toward automated textual moral reasoning.  
\end{abstract}

\section{Introduction}
The relation of morality and computational intelligence has been a topic of philosophical inquiry among the early pioneers of artificial intelligence~\cite{turing1950computing,wiener1960some}. Recent advances in artificial intelligence in areas such as autonomous driving makes it desirable for intelligent systems to develop moral awareness (e.g.,~\citealp{awad2018moral}). 
The ability to reason about everyday moral situations is one key aspect that distinguishes humans from machines, and language provides a natural conduit for generic moral inference. Here we explore methodologies for automated inference of moral vignettes from textual input.


There exist multiple schools of thought on human morality. Some theories present morality as a form of deliberate reasoning, while others assert that moral judgment is driven largely by emotion. Recent research from moral psychology has suggested that morality depends on five to six foundational dimensions shared across cultures~\cite{haidt2004intuitive,haidt2007moral,graham2013moral}. As such, morality may be viewed as intuitive judgement that nevertheless depends on fine-grained categorical inference. Our emphasis  here departs from the theoretical debate and instead focuses on examining the possibility of machine inference for daily moral situations. 

Given a generic  vignette such as ``if you tell a secret someone told you'', we ask whether the human-judged moral category of the vignette (e.g., {\it loyalty} in this case) can be automatically inferred from text alone. One important aspect we explore here is whether the moral vignettes can be represented effectively, and in particular we examine several alternative  representations including the recent contextualized language models (e.g.,~\citealp{reimers-2019-sentence-bert}). Another avenue we explore is whether these representations can be generalized accurately onto fine-grained moral categories (e.g., fairness, care, loyalty) above and beyond the binary distinction between right and wrong (e.g.,~\citealp{schramowski2019bert}), given only a limited number of human examples as training data. 




\section{Related work}



A few disparate ideologies concerning morality have prevailed in philosophy. Immanuel Kant proposed that morality is grounded in reasoning, and that reflecting on principles is how moral judgment ought to be made~\cite{kant2005moral}. To the contrary, Hume argued that reasoning was second to emotion: It is feelings and emotions that primarily guide us in our moral judgment~\cite{hume1902enquiries}. The recent advent of the Moral Foundations Theory (MFT)~\cite{graham2013moral}  presents a modular view of morality based on a core set of dimensions or foundations. Examples of such foundations include care, fairness, loyalty, authority, purity, etc. The moral foundations dictionary (MFD)~\cite{graham2009liberals} was created as a linguistic resource to bridge moral psychology and natural language processing, where key lexical terms were listed under each moral foundation. The MFT and MFD have been the basis of several studies in natural language processing, which have typically focused on predicting moral sentiment from social media text \cite{mooijman2018moralization,lin2018acquiring}. Other lines of work have explored moral rhetoric in political discourse \cite{garten2016morality}, inference of moral sentiment change from diachronic word embeddings \cite{xie2020text}, classification of moral sentiment from tweets \cite{hoover2019moral}, and analyses of moral bias toward right or wrong using contextualized language models SBERT~\cite{schramowski2019bert}. Our study extends this line of research by exploring textual inference of moral vignettes that often entail complex world knowledge, and we contribute a set of empirical data collected from  moral psychology experiments which are under-represented in the existing natural language processing studies of morality.


\section{Data}
We collect text-based moral vignettes, each of which refers to a scenario in the form of a short text, e.g., ``A guy cheats his family out of their money and property" \cite{clifford}.
Moral vignettes have traditionally been presented as stimuli to study the bases of human moral judgment in moral psychology.
We collect vignettes from  the  following three independent sources, with their characteristics summarized in Table \ref{tab:vignettes}:
\begin{itemize}
    \item \textbf{Set 1 -- Chadwick.}
    \citet{chadwick} present a rich source of pretested behaviours, each corresponding to either a positive or negative example of the traits of honesty, loyalty, friendliness, charitableness.
    \item \textbf{Set 2 -- McCurrie.}
     \citet{mccurrie} provide video clips of moral stimuli that depict violations or the five vice dimensions of the Moral Foundations Theory: harm, unfairness or cheating, disloyalty, anti-authority,  and (sexual) impurity. We use the textual description of the content from each video clip. Videos were sourced from YouTube and validated through human ratings.
    \item \textbf{Set 3 -- Clifford.}
    \citet{clifford} present a set of moral violations, each of which is an infringement on one of the five Moral Foundations. Vignettes were written to incorporate a varied content and minimize overlap among the foundations. The vignettes were validated through factor analysis and human participants.
\end{itemize}

\begin{table}[]
\caption{Key characteristics of the moral vignette sets.}
\label{tab:vignettes}
\begin{tabular}{lrlp{2.8cm}}
\hline
Set   & Size & Polarity & Categories \\ \hline
Chadwick & 500  & +ve/-ve     & honesty, loyalty, friendliness, charitableness, cooperativeness \\
McCurrie &  69   & -ve & MFT foundations                                                 \\
Clifford & 132  & -ve & MFT foundations                                                 
\end{tabular}
\end{table}

For both vignette datasets associated with MFT, we considered vignettes from the five categories and excluded the sixth category, liberty. This is in an effort to be consistent with the MFD v1.0 used in our models that does not contain words pertaining to liberty.

The dataset is available \href{https://github.com/xiejxie/moral-sentiment-prediction}{here}.

\section{Methodology}
We explore how various representations of moral vignettes contribute to effective inference of human moral judgment.
For each dataset, we transform each textual scenario into each of the five representations. We choose these representations to reflect the most relevant aspects of human moral judgment grounded in existing theories on moral reasoning and emotion, and state-of-the-art language models. Standard classification methods are applied in 5-fold cross-validation of the resulting representations.
Classes are determined by the trait categorizations of the vignettes summarized in Table \ref{tab:vignettes}.
We evaluate model performance across all permutations of the datasets ($\times$ 3), representational schemes ($\times$ 5), and classification methods ($\times$ 4).
\subsection{Representations}
\begin{itemize}
    \item \textbf{Contextual Embeddings.}
    Contextual embeddings capture meanings for words as a function of their context.
    We use a pretrained sentence-based BERT model to embed each vignette \cite{reimers-2019-sentence-bert}.
    \item \textbf{Average Embeddings.}
    As a baseline to contextual embeddings, we take the average of word embeddings in a vignette.
    In this approach, we consider scenarios as a whole, drawing information from its  background semantic knowledge.
    Embeddings are pretrained using GloVe \cite{pennington2014glove}.
    \item \textbf{Verb Embeddings.}
    With verb embeddings, we extract the root verb of a vignette and perform classification based solely on the root.
    The root is determined through dependency parsing \cite{manning-EtAl:2014:P14-5}.
    We also use GloVe vectors to produce embeddings for the verb.
    Deontological ethics \cite{kant2005moral, alexander2007deontological} in particular is fundamentally concerned with the inherent rightness and wrongness of an action.
    Due to the high variability in context, it is feasible that the verb is strong enough signal on its own.
    For instance, ``to kill" carries a heavy negative association.
    In the majority of cases, it is not necessary to know who or what is the direct object.
    \item \textbf{Moral Sentiment.}
    We extend previous work \cite{garten2018dictionaries, xie2020text} to create a representation grounded in the MFT.
    We compute centroids $c_f$ for each foundation $f$.
    The centroid is the mean of all word embeddings $w_{f}$ in $f$: 
        $c_f = \frac{1}{N}\sum w_{f}$.
    The euclidean distance to each centroid for each word vector $w_v$ in a vignette is then calculated.
    The mean of these distances for each $f$ is taken as a feature $e_f$: $e_f = ||w-c_f||_2$.
    \item \textbf{Emotion.}
    As theorized by \citet{hume1902enquiries}, emotion may play an integral role in how humans make moral judgment.
    We consider representations based on the affective ratings of individual ratings.
    We use the valence, arousal and dominance mean scores from \cite{warriner2013norms}.
    Scores are then averaged across all words in a vignette.
\end{itemize}
\subsection{Classifiers}
We use a range of standard models for classification, exploiting various aspects of the data.
Gaussian Naive Bayes (NB) and Regression (LR) classifiers are used as standard methods.
Also considered is a $k$ Nearest Neighbors (kNN) model, which makes predictions from local density. Although we explored at various values for $k$, results shown are for $k$ = 5, which has proven robust against all datasets.
Additionally, SVM classifiers (SVC)  amenable to high-dimensional data \cite{suykens1999least} are also considered.

\section{Results}
\begin{figure*}
\centering
\caption{Projection of all vignette contextual embeddings onto a 2-D space. Example points are marked for 
each dataset and category. In the Chadwick (binary) plot, ``charitable", ``cooperative", ``friendly", ``honest" and ``loyal" labels are grouped as Positive.}
\label{fig:tsne}
\begin{subfigure}
    \centering
    \includegraphics[height=0.4\linewidth]{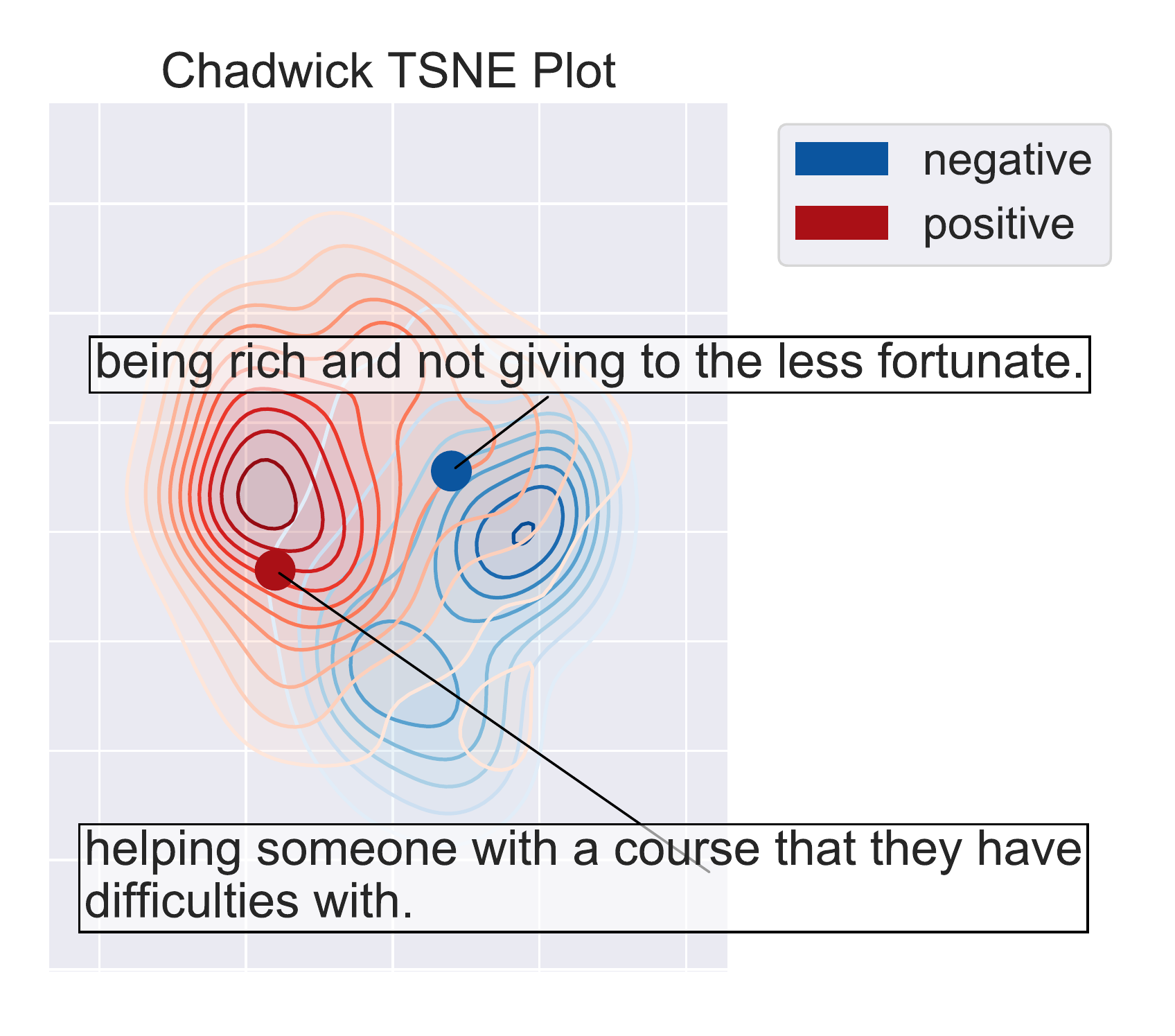}
\end{subfigure}
\begin{subfigure}
    \centering
    \includegraphics[height=0.4\linewidth]{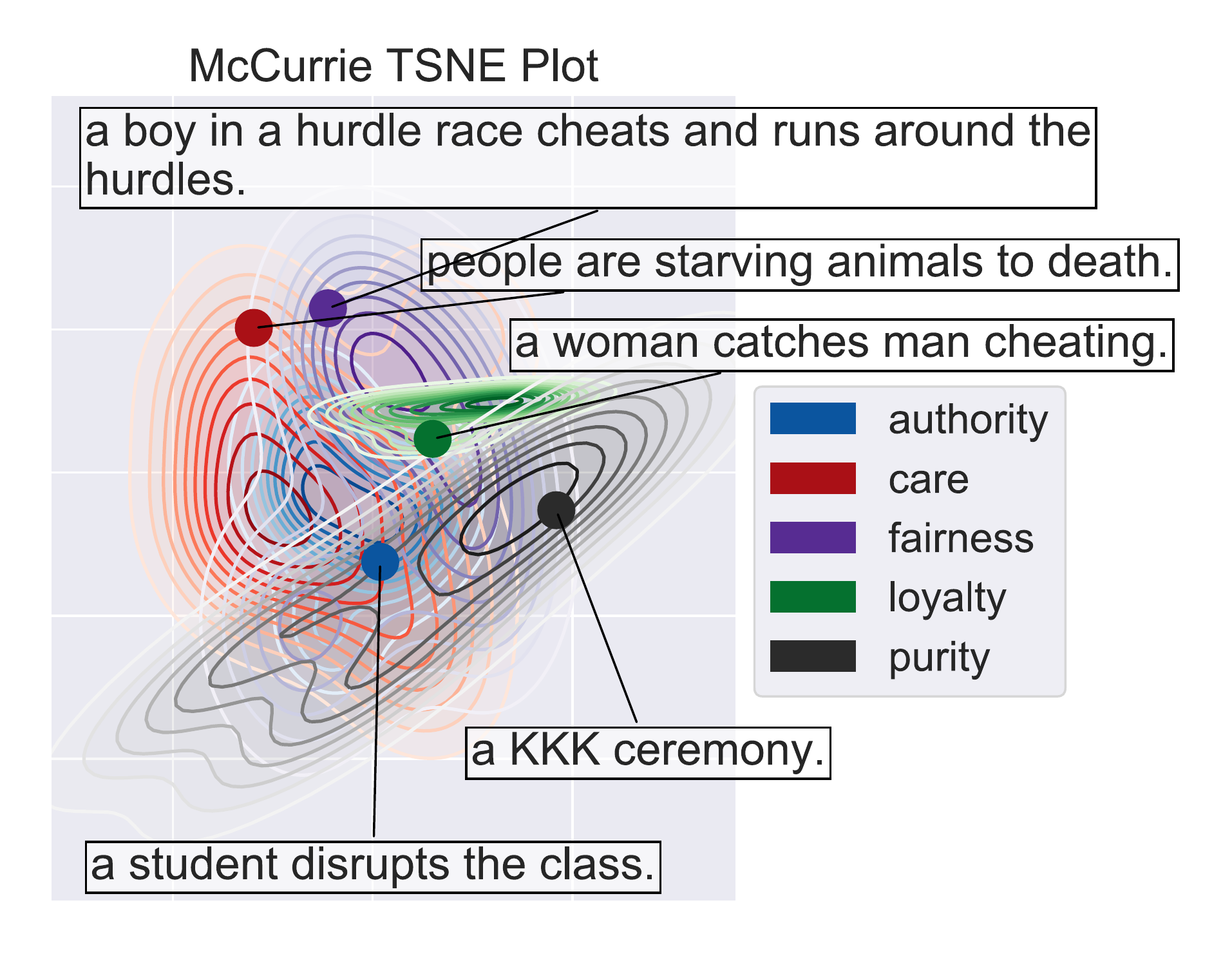}
\end{subfigure}
\end{figure*}

We summarize the results of model evaluation against human judgement in Table \ref{tab:feature-breakdown}.
Given that random chance is 10\% for all datasets, the majority of models correlate  with human judgment substantially above chance. Despite limited training examples, the contextual embeddings yield the best overall performance across the 3 datasets and among the 5 representations, and this advantage is robust across the 4 classification methods. This may be a direct consequence of the rich semantic information encoded into the contextual models, as well as its ability to capture multiple senses of a word \cite{hewitt2019structural}. Logistic regression produces the highest accuracy in most cases. A visual representation of the contextual embeddings is shown in Figure \ref{fig:tsne}, which displays the density of moral categories for the vignettes, projected onto two dimensions using TSNE \cite{maaten2008visualizing}. 
We observe a clean delineation in the Chadwick set, suggesting a division between vignettes that are morally acceptable versus not. We also observe categorical structure in the McCurrie set that illustrates a fine-grained moral inferential capacity.

\begin{table*}[]
\centering
\caption{Performance by representation and classification method. Model accuracy is averaged over 5-fold cross-validation.}
\label{tab:feature-breakdown}
\begin{tabular}{llccccc}
\hline
Dataset  & Representation   & NB      & kNN     & LR      & SVC     & Mean accuracy (SD) \\ \hline
Chadwick & Context. Embed. (SBERT)       & 44.80\% & 46.20\% & 51.80\% & 50.00\% & \textbf{48.20\% (2.81\%)}\\
Chadwick & Avg. Embed. (GloVe)      & 16.80\% & 14.00\% & 13.20\% & 6.80\%  & 12.70\% (3.66\%)      \\
Chadwick & Verb Embed. (GloVe) & 28.67\% & 27.45\% & 34.28\% & 30.67\% & 30.27\% (2.58\%) \\
Chadwick & Moral Sentiment & 11.40\% & 11.00\% & 8.40\%  & 7.00\%  & 9.45\% (1.82\%) \\ 
Chadwick & Emotion    & 11.80\% & 9.40\%  & 8.00\%  & 12.60\% & 10.45\% (1.84\%)\\\hline
Clifford & Context. Embed. (SBERT)      & 58.68\% & 51.68\% & 65.79\% & 54.63\% & \textbf{57.70\% (5.29\%)} \\
Clifford & Avg. Embed. (GloVe)      & 24.37\% & 28.37\% & 32.32\% & 32.32\% & 29.34\% (3.29\%) \\
Clifford & Verb Embed. (GloVe) & 47.47\% & 30.21\% & 58.58\% & 46.53\% & 45.70\% (10.11\%) \\
Clifford & Moral Sentiment  & 25.32\% & 38.11\% & 32.26\% & 32.26\% & 31.99\% (4.53\%) \\ 
Clifford & Emotion    & 18.11\% & 15.16\% & 23.32\% & 27.26\% & 20.96\% (4.66\%) \\\hline
McCurrie & Context. Embed. (SBERT)      & 47.95\% & 38.21\% & 49.23\% & 42.82\% & \textbf{44.55\% (4.38\%)} \\
McCurrie & Avg. Embed. (GloVe)     & 35.00\% & 28.33\% & 37.95\% & 37.95\% & 34.81\% (3.93\%) \\
McCurrie & Verb Embed. (GloVe) & 36.54\% & 41.41\% & 47.56\% & 41.15\% & 41.67\% (3.92\%) \\
McCurrie & Moral Sentiment  & 28.33\% & 31.67\% & 37.95\% & 37.95\% & 33.97\% (4.15\%)\\
McCurrie & Emotion    & 36.28\% & 31.67\% & 37.95\% & 37.82\% & 35.93\% (2.55\%) \\
\end{tabular}
\end{table*}


\begin{table*}[]
\caption{Errors committed by the Contextual Embeddings representation under the logistic regression classifier.}
\label{tab:bert-errors}
\begin{tabular}{lp{10.0cm}ll}
\hline
Dataset  & Vignette                                                                 & Prediction & Truth        \\ \hline
Chadwick & Holding back snide comments you have for a team member.                  & Unfriendly & Cooperative  \\
Chadwick & Helping someone with a course that they have difficulties with.          & Charitable & Friendly     \\
Chadwick & Not offering service to someone who isn't dressed up to par.              & Charitable & Uncharitable \\ \hline
McCurrie & A guy cheats his family out of their money and property.                 & Fairness   & Loyalty      \\
McCurrie & A man takes drugs on a bus.                                              & Care       & Purity       \\
McCurrie & A basketball player yells at his coaches.                                & Care       & Loyalty      \\ \hline
Clifford & A soccer player pretends to be seriously fouled by an opposing player.   & Authority  & Fairness     \\
Clifford & A man leaves his family business to go work for their main competitor.   & Care       & Loyalty      \\
Clifford & A Hollywood star agrees with a foreign dictator's denunciation of the US. & Authority  & Loyalty     
\end{tabular}
\end{table*}


\subsection{Error interpretation} 

To examine the limitations of the contextual approach, we extract errors in Table \ref{tab:bert-errors} under the best performing logistic regression model. In several examples, the model is misled by words that are strongly associated with alternative categories. For instance, ``cheating" is in violation of fairness, but the vignette is categorized as an infraction of loyalty (row 4). Another common error is the handling of negation (row 3), only some of which were predicted correctly.



\section {Conclusion}

We present a first exploration of textual inference for moral vignettes and demonstrate the effectiveness and limitations of predicting human  categorization of everyday moral situations under limited training data.
We show that an approach based on contextualized representation offers a superior performance over alternative representations based on existing theories of morality and non-contextualized word embeddings. Future work should explore common knowledge and reasoning in moral inference beyond semantic information, and our current study serves as a stepping stone toward  automated moral reasoning from text.


\bibliographystyle{acl_natbib}
\bibliography{references}




\end{document}